\documentclass{article}

\PassOptionsToPackage{numbers, compress}{natbib}

\usepackage[preprint]{neurips_2022}

\usepackage[utf8]{inputenc} 
\usepackage[T1]{fontenc}    
\usepackage{hyperref}       
\usepackage{url}            
\usepackage{booktabs}       
\usepackage{amsfonts}       
\usepackage{nicefrac}       
\usepackage{microtype}      
\usepackage{xcolor}         

\usepackage{amsmath}
\usepackage{graphicx}
\usepackage{multirow}
\usepackage{pifont}
\newcommand{\cmark}{\ding{51}}%
\newcommand{\xmark}{\ding{55}}%
\newcommand{\boldtheta}{\boldsymbol \theta}

\title{Text Embeddings by Weakly-Supervised \\ Contrastive Pre-training}

\author{Liang Wang,~Nan Yang,~Xiaolong Huang,~Binxing Jiao\\
~\textbf{Linjun Yang},~\textbf{Daxin Jiang},~\textbf{Rangan Majumder},~\textbf{Furu Wei}\\
Microsoft Corporation \\
\url{https://github.com/microsoft/unilm} \\}

\begin{document}

\maketitle

\begin{abstract}
    This paper presents E5~\footnote{E5: \textbf{E}mb\textbf{E}ddings from bidir\textbf{E}ctional \textbf{E}ncoder r\textbf{E}presentations},
    a family of state-of-the-art text embeddings that transfer well to a wide range of tasks.
    The model is trained in a contrastive manner
    with weak supervision signals from our curated large-scale text pair dataset (called CCPairs).
    E5 can be readily used as a general-purpose embedding model for any tasks requiring a single-vector representation of texts such as retrieval, clustering, and classification, achieving strong performance in both zero-shot and fine-tuned settings.
    We conduct extensive evaluations on $56$ datasets from the BEIR and MTEB benchmarks.
    For zero-shot settings, E5 is the first model that outperforms the strong BM25 baseline on the BEIR retrieval benchmark without using any labeled data.
    When fine-tuned, E5 obtains the best results on the MTEB benchmark, beating existing embedding models with $40\times$ more parameters.
\end{abstract}

\section{Introduction}
Text embeddings are low-dimensional vector representations for arbitrary-length texts
and play key roles in many NLP tasks such as large-scale retrieval.
Compared to the high-dimensional and sparse representations like TF-IDF,
text embeddings have the potential to overcome the lexical mismatch issue
and facilitate efficient retrieval and matching between texts.
It also offers a versatile interface easily consumable by downstream applications.

While pre-trained language models such as
BERT ~\citep{Devlin2019BERTPO} and GPT ~\citep{NEURIPS2020_1457c0d6}
can produce transferrable text representations,
they are not ideal for tasks such as retrieval and text matching
where a single-vector embedding of texts is more desired
due to its efficiency and versatility.
To obtain better text embeddings, contrastive learning is
often the go-to framework to
enhance the sequence-level representations from text pairs.
Along this line of research,
some works are geared towards learning task-specific embeddings.
For example, GTR ~\citep{Ni2021LargeDE} and Sentence-T5 ~\citep{ni2022sentence} fine-tune pre-trained models
with supervised datasets to learn embeddings customized for passage retrieval and semantic textual similarity, respectively.
Other works learn unsupervised embeddings
from automatically constructed text pairs.
Typical methods to construct text pairs include Inverse Close Task (ICT) ~\citep{Chang2020PretrainingTF}, random cropping ~\citep{Izacard2021TowardsUD} and neighboring text spans ~\citep{Neelakantan2022TextAC}, etc. While such synthetic data are of unlimited quantity, they are often poor in quality and the resulted embeddings fail to match the performance of the classic BM25 baseline without further fine-tuning \citep{Muennighoff2022MTEBMT}.

In this work, we learn a high-quality general-purpose text embedding termed E5, \textbf{E}mb\textbf{E}ddings from bidir\textbf{E}ctional \textbf{E}ncoder r\textbf{E}presentations.
E5 aims to provide strong off-the-shelf text embeddings suitable for any tasks requiring single-vector representations in both zero-shot or fine-tuned settings.
To achieve this goal,
instead of relying on limited labeled data or low-quality synthetic text pairs,
we contrastively train E5 embeddings from CCPairs, a curated web-scale text pair dataset containing heterogeneous training signals.
We construct the CCPairs dataset by combining various semi-structured data sources such as CommunityQA, Common Crawl and Scientific papers, and perform aggressive filtering with a consistency-based filter ~\citep{Dai2022PromptagatorFD} to improve data quality.
We choose a simple contrastive learning recipe using in-batch negatives with a large batch-size to train our model.
Extensive experiments on both BEIR and MTEB benchmarks demonstrate the effectiveness of the proposed method.
On the BEIR zero-shot retrieval benchmark ~\citep{Thakur2021BEIRAH},
E5 is the first model to outperform the strong BM25 baseline without using any labeled data.
When fine-tuned on labeled datasets,
the performance can be further improved.
Results on $56$ datasets from the recently introduced MTEB benchmark ~\citep{Muennighoff2022MTEBMT} show that
our E5$_\text{base}$ is competitive against GTR$_\text{xxl}$ and Sentence-T5$_\text{xxl}$,
which have $40\times$ more parameters.

\section{Related Work}

There have been long-lasting interests in transforming texts into low-dimensional dense embeddings.
Early works include Latent Semantic Indexing (LSA) ~\citep{deerwester1990indexing} and
Latent Dirichlet Allocation (LDA) ~\citep{blei2003latent}.
LSA utilizes the decomposition of a word-document co-occurrence matrix to generate document embeddings,
while LDA adopts probabilistic graphical models to learn topic distributions.
~\citeauthor{arora2019simple} show that a simple weighted average of word vectors ~\citep{Mikolov2013EfficientEO}
can be a strong baseline for sentence embeddings.

With the development of pre-trained language models ~\citep{Devlin2019BERTPO,Liu2019RoBERTaAR,raffel2020exploring}
and large-scale labeled datasets such as SNLI ~\citep{Bowman2015ALA} and MS-MARCO ~\citep{Campos2016MSMA},
methods like Sentence-BERT ~\citep{Reimers2019SentenceBERTSE}, SimCSE ~\citep{Gao2021SimCSESC},
Sentence-T5 ~\citep{ni2022sentence} and SGPT ~\citep{Muennighoff2022SGPTGS}
directly fine-tune language models to output continuous embeddings.
Most research focuses on short texts and thus uses the term "sentence embeddings".
For long documents,
it remains an open research question whether fixed-length embeddings can encode all the information.
Contrastive loss popularized by SimCLR ~\citep{Chen2020ASF} turns out to be more effective
than classification-based losses ~\citep{Reimers2019SentenceBERTSE, Conneau2017SupervisedLO} for embeddings.
LaBSE ~\citep{feng2022language}, LASER ~\citep{artetxe2019massively} and CLIP ~\citep{Radford2021LearningTV}
further extend to multilingual and multi-modal scenarios using parallel sentences and image-text pairs.

Another direction is to design self-supervised pre-training tasks for text matching and retrieval.
~\citep{Chang2020PretrainingTF} proposes the well-known inverse cloze task (ICT),
where a random sentence within a passage is chosen as a pseudo-query and
the rest is treated as a positive sample.
However,
Contriever ~\citep{Izacard2021TowardsUD} shows that random cropping with data augmentation
is more effective than ICT on a range of zero-shot information retrieval tasks.
OpenAI text embeddings ~\citep{Neelakantan2022TextAC} use neighboring texts as positives and
scale up the model size to $175$B.
~\citet{Ouz2022DomainmatchedPT} performs domain-matched pre-training to improve in-domain results.
SPAR ~\citep{chen2021salient} trains a dense retriever by treating BM25 as a teacher model.
Although the aforementioned approaches can easily obtain abundant supervision signals,
such synthetic data tend to be of low quality.
Results on the BEIR benchmark ~\citep{Thakur2021BEIRAH} show
they struggle to match the performance of BM25
if not further fine-tuned on labeled datasets.

Evaluation and interpretation of text embeddings are also non-trivial.
Most benchmarks measure the embedding quality through downstream task performances.
For example,
SentEval ~\citep{Conneau2018SentEvalAE} uses linear probing and
a collection of semantic textual similarity (STS) datasets,
while the BEIR benchmark ~\citep{Thakur2021BEIRAH} focuses on zero-shot information retrieval scenarios.
The recently introduced MTEB benchmark ~\citep{Muennighoff2022MTEBMT}
combines $56$ datasets spanning across $8$ tasks and $112$ languages.
Experiments show no model can achieve state-of-the-art results on all embedding tasks yet.
In this paper,
we do not use the SentEval toolkit since its linear probing setup depends on the optimization hyperparameters.

Most closely related to our work is a series of community efforts
by \emph{sentence-transformers} ~\footnote{\url{https://github.com/UKPLab/sentence-transformers}}
to train embeddings with a collection of labeled and automatically collected datasets.
In this paper,
we show that it is possible to train high-quality embeddings using self-supervised pre-training only.
In terms of benchmark results,
our model can achieve superior performance when fine-tuned on less labeled data.

\section{CCPairs: A Large Collection of Text Pair Dataset}

\begin{figure}[ht]
\begin{center}
 \includegraphics[width=1.05\linewidth]{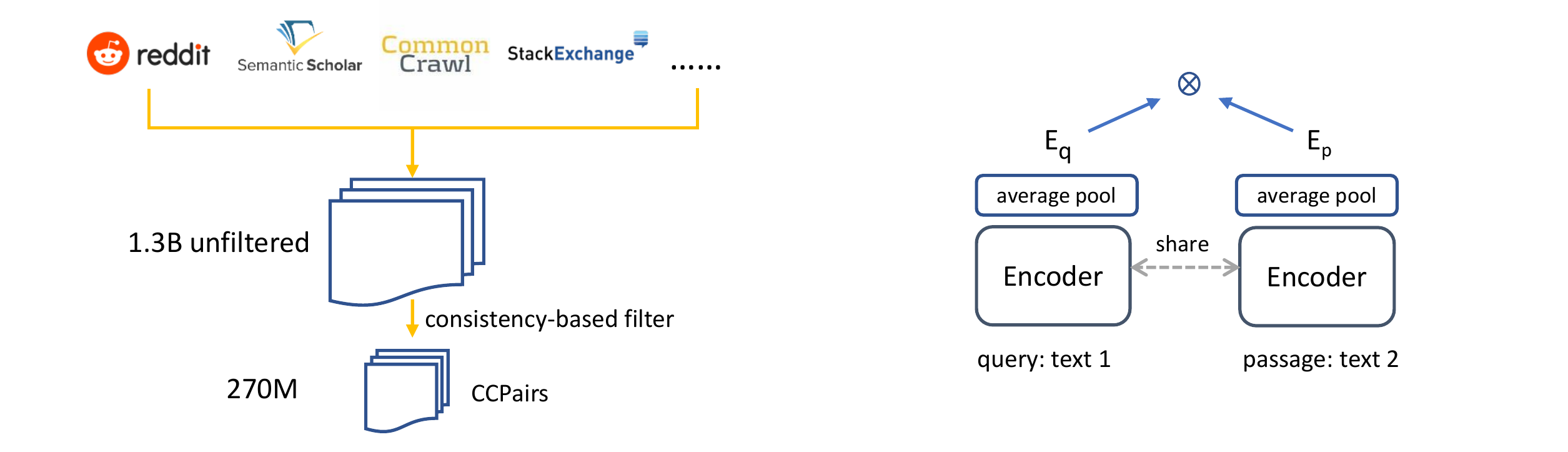}
 \caption{Overview of our data curation pipeline and model architecture.}
 \label{fig:overview}
\end{center}
\end{figure}

The quality and diversity of the data is crucial for training general-purpose text embeddings.
In this work, we mine and assemble CCPairs, a large high-quality text pair dataset from web sources
which provide diverse training signals transferring well to a wide range of tasks.

\noindent
\textbf{Harvesting semi-structured data sources }
Large-scale high-quality datasets
like C4 ~\citep{raffel2020exploring} and CCMatrix ~\citep{Schwenk2021CCMatrixMB}
are vital for the success of language model pre-training and machine translation.
For learning text embeddings,
existing works either utilize small-scale human-annotated data
such as NLI ~\citep{Gao2021SimCSESC} and MS-MARCO ~\citep{Campos2016MSMA}
or adopt heuristics such as random cropping ~\citep{Izacard2021TowardsUD}
to obtain large-scale but very noisy supervision signals.

Instead,
we curate a text pair dataset CCPairs (\textbf{C}olossal \textbf{C}lean text \textbf{Pairs})
by harvesting heterogeneous semi-structured data sources.
Let ($q$, $p$) denote a text pair
consisting of a query $q$ and a passage $p$.
Here we use ``\emph{passage}'' to denote word sequences of arbitrary length,
which can be a short sentence, a paragraph, or a long document.
Our dataset includes (post, comment) pairs from Reddit ~\footnote{\url{https://files.pushshift.io/reddit/}},
(question, upvoted answer) pairs from Stackexchange ~\footnote{\url{https://archive.org/details/stackexchange}},
(entity name + section title, passage) pairs from English Wikipedia,
(title, abstract) and citation pairs from Scientific papers ~\citep{lo2020s2orc},
and (title, passage) pairs from Common Crawl ~\footnote{\url{https://commoncrawl.org/}}
web pages and various News sources.

We only include data sources that can be automatically mined,
and some subsets are directly reused from existing datasets.
Simple heuristic rules are applied to filter data from Reddit and Common Crawl.
For example,
we remove Reddit comments that are either too long ($> 4096$ characters)
or receive score less than $1$,
and remove passages from web pages with high perplexity ~\citep{wenzek2019ccnet}.
After preliminary filtering,
we end up with $\sim1.3$ billion text pairs,
most of which come from Reddit and Common Crawl.
For more details and examples,
please refer to Appendix ~\ref{sec:app_data_details}.

\noindent
\textbf{Consistency-based filter }
To further improve data quality and make training costs manageable,
we propose a consistency-based data filtering technique:
a model is first trained on the $1.3$B noisy text pairs,
and then used to rank each pair against a pool of $1$ million random passages.
A text pair is kept only if it falls in the top-$k$ ranked lists.
In other words,
the model's prediction should be consistent with the training labels.
Here we set $k = 2$ based on manual inspection of data quality.
After this step,
we end up with $\sim270$M text pairs for contrastive pre-training.

The intuition for this technique comes from the
memorization behaviors of neural networks ~\citep{Feldman2020WhatNN}:
when trained on noisy datasets,
neural networks tend to memorize the clean labels first
and then gradually overfit the noisy labels.
Similar techniques ~\citep{nguyen2019self,Dai2022PromptagatorFD,Han2018CoteachingRT}
have been widely used for removing dataset noises.
It is also possible to apply this filter iteratively,
we will leave it for future work.

\section{Method}

Our embeddings can be trained with only unlabeled text pairs from CCPairs with contrastive pre-training. A second-stage fine-tuning on small, high-quality labeled datasets can be performed to further boost the quality of the resulted embeddings. See Figure~\ref{fig:overview} for an overview.

\subsection{Contrastive Pre-training with Unlabeled Data}

Contrastive pre-training aims to distinguish the relevant
text pairs from other irrelevant or negative pairs.
Given a collection of text pairs $\{(q_i, p_i)\}_{i=1}^n$,
we assign a list of negative passages $\{p_{ij}^-\}_{j=1}^m$ for the $i$-th example.
Then the InfoNCE contrastive loss ~\citep{Chen2020ASF} is as follows:
\begin{equation} \label{equ:infonce}
    \min\ \ L_\text{cont} = -\frac{1}{n} \sum_i \log \frac{\text{e}^{s_{\boldtheta} (q_i,p_i)}}{\text{e}^{s_{\boldtheta} (q_i,p_i)}+\sum_j \text{e}^{s_{\boldtheta} (q_i,p_{ij}^-)} }
\end{equation}
where $s_{\boldtheta}(q,p)$ is a scoring function between
query $q$ and passage $p$ parameterized by $\boldtheta$.
Following the popular biencoder architecture,
we use a pre-trained Transformer encoder and average pooling over the output layer
to get fixed-size text embeddings $\mathbf{E}_q$ and $\mathbf{E}_p$.
The score is the cosine similarity scaled by a temperature hyperparameter $\tau$:
\begin{equation}
    s_{\boldtheta}(p, q) = \text{cos(}\mathbf{E}_q\text{, } \mathbf{E}_p\text{)}\ /\ \tau
\end{equation}
Where $\tau$ is set to $0.01$ in our experiments by default.
We use a shared encoder for all input texts
and break the symmetry by adding two prefix identifiers \emph{``query:''} and \emph{``passage:''}
to $q$ and $d$ respectively.
For some data sources such as citation pairs,
it is not obvious which side should be the query,
we randomly choose one for simplicity.
Such an asymmetric design turns out to be important for some retrieval tasks
where there exist paraphrases of the query in the target corpus.

Another critical issue for contrastive training is
how to select the negative samples.
Here we choose to use the in-batch negatives ~\citep{Chen2020ASF},
where the passages from other pairs in a batch serve as negative samples.
We find that this simple strategy enables more stable training
and outperforms methods such as MoCo ~\citep{He2020MomentumCF}
when the batch size is sufficiently large.

\subsection{Fine-tuning with Labeled Data}

While contrastive pre-training on the CCPairs provides a solid foundation for general-purpose embeddings, further training on labeled data can inject human knowledge into the model to boost the performance.
Although these datasets are small,
existing works ~\citep{Ni2021LargeDE,ni2022sentence} have shown
that supervised fine-tuning leads to consistent performance gains.
In this paper,
we choose to further train with a combination of 3 datasets:
NLI~\footnote{The version released by SimCSE.} (Natural Language Inference),
MS-MARCO passage ranking dataset ~\citep{Campos2016MSMA},
and NQ (Natural Questions) dataset ~\citep{Karpukhin2020DensePR,Kwiatkowski2019NaturalQA}.
Empirically,
tasks like STS (Semantic Textual Similarity) and linear probing benefit from NLI data,
while MS-MARCO and NQ datasets transfer well to retrieval tasks.

Building on the practices of training state-of-the-art dense retrievers ~\citep{ren2021rocketqav2,Wang2022SimLMPW},
we use mined hard negatives and knowledge distillation from a cross-encoder (CE) teacher model
for the MS-MARCO and NQ datasets.
For the NLI dataset,
contradiction sentences are regarded as hard negatives.
The loss function is a linear interpolation between
contrastive loss $L_\text{cont}$ for hard labels
and KL divergence $D_\text{KL}$ for distilling soft labels from the teacher model.
\begin{equation}
    \min\ \ D_{\text{KL}}(p_\text{ce},p_\text{stu}) + \alpha  L_\text{cont}
\end{equation}
Where $p_\text{ce}$ and $p_\text{stu}$ are the probabilities
from the cross-encoder teacher model and our student model.
$\alpha$ is a hyperparameter to balance the two loss functions.
$L_\text{cont}$ is the same as in Equation ~\ref{equ:infonce}.

\subsection{Applications to Text Embedding Tasks} \label{sec:transfer}

After the above two steps, we obtain high-quality text embeddings transferring well to a wide range of tasks
without fine-tuning the model parameters.
Combined with techniques like approximate nearest neighbor search,
embeddings provide a scalable and efficient solution for applications like web search.
Here we briefly illustrate several use cases of our text embeddings.

\noindent
\textbf{Zero-shot Retrieval }
First,
the passage embeddings for the target corpus are computed and indexed offline.
Then for each query,
we compute its query embedding and return the top-$k$ ranked lists
from the corpus based on cosine similarity.

\noindent
\textbf{Few-shot Text Classification }
A linear classifier is trained on top of the frozen embeddings
with a few labeled examples.
Different tasks only need to train and save the parameters of the classification heads.
It can be seen as a particular form of parameter-efficient learning ~\citep{Houlsby2019ParameterEfficientTL}.

\noindent
\textbf{Zero-shot Text Classification }
The input and label texts are converted to sentences based on
manually written prompt templates.
The predicted label is the one closest to the input text in the embedding space.
Take the sentiment classification of movie reviews as an example,
with the original input ``\emph{I enjoy watching it}'',
the label text is ``\emph{it is an example of terrible/great movie review}''
and the input text becomes ``\emph{movie review: I enjoy watching it}''.

\noindent
\textbf{Semantic Textual Similarity }
Given two text embeddings,
we use the cosine function to measure their semantic similarity.
Since the absolute similarity scores do not enable an easy interpretation,
the evaluation is usually based on rank correlation coefficients.

\noindent
\textbf{Text Clustering }
Standard clustering algorithms such as k-means can be applied straightforwardly.
Texts belonging to the same category are expected to be close in the embedding space.

For tasks other than zero-shot text classification and retrieval,
we use the query embeddings by default.

\section{Experiments}

\subsection{Pre-training and Fine-tuning Configurations}

\noindent
\textbf{Pre-training }
We pre-train on our proposed text pair dataset for three model sizes:
E5$_\text{small}$, E5$_\text{base}$ and E5$_\text{large}$
initialized from MiniLM ~\citep{wang2021minilmv2},
\emph{bert-base-uncased}, and \emph{bert-large-uncased-whole-word-masking} respectively.
The batch size is set to a large value of $32,768$ to increase the number of negatives.
The learning rate is \{$3, 2, 1$\}$\times10^{-4}$ for the \{small, base, large\} models,
with linear decay and the first $1,000$ steps for warmup.
We pre-train for $20k$ steps in total with AdamW optimizer,
which is approximately $2.5$ epochs over the dataset.
It takes \{$16, 32, 64$\} V100 GPUs and \{$1, 1, 2$\} days for the \{small, base, large\} models.
To improve training efficiency and reduce GPU memory usage,
we adopt mixed precision training and gradient checkpointing.

\noindent
\textbf{Fine-tuning }
is performed on the concatenation of $3$ datasets:
MS-MARCO passage ranking ~\citep{Campos2016MSMA},
NQ ~\citep{Kwiatkowski2019NaturalQA,Karpukhin2020DensePR},
and NLI ~\citep{Gao2021SimCSESC} datasets.
We reuse the mined hard negatives and re-ranker scores from SimLM ~\citep{Wang2022SimLMPW}
for the first two datasets.
Models are fine-tuned for $3$ epochs with batch size $256$ on $8$ GPUs.
Learning rate is \{$3, 2, 1$\}$\times10^{-5}$ for the \{small, base, large\} models
with $400$ steps warmup.
For each example,
we use $7$ hard negatives.
Since the NLI dataset only has $1$ hard negative for each example,
$6$ sentences are randomly sampled from the entire corpus.

We use E5-PT to denote models with contrastive pre-training only.
More implementation details can be found in Appendix ~\ref{sec:app_impl}.

\subsection{Evaluation Datasets}

\noindent
\textbf{BEIR Benchmark ~\citep{Thakur2021BEIRAH} }
is a collection of $19$ information retrieval datasets,
ranging across ad-hoc web search, question answering,
fact verification and duplicate question retrieval, etc.
We evaluate the $15$ datasets that provide public downloads.
The main metric is nDCG@$10$.

\noindent
\textbf{MTEB Benchmark ~\citep{Muennighoff2022MTEBMT} }
is recently proposed for benchmarking massive text embedding tasks.
Though MTEB is multilingual due to the inclusion of bitext mining datasets,
most datasets are still only available in English.
In this paper,
we evaluate the English subsets,
which have $56$ datasets spanning across $6$ categories:
Classification (Class.), Clustering (Clust.),
Pair Classification (PairClass.), Rerank,
Retrieval (Retr.), STS, and Summarization (Summ.).
The evaluation metrics are accuracy, v-measure,
average precision, MAP, nDCG@10, and Spearman coefficients,
respectively.
Please refer to the MTEB paper for details.

\subsection{Results on BEIR benchmark}

\begin{table}[ht]
\centering
\caption{Unsupervised methods on the BEIR benchmark (nDCG@$10$).
For SimCSE, we report results with BERT$_\text{base}$.
cpt$_{\text{300M}}$ ~\citep{Neelakantan2022TextAC} is only available through paid API
and evaluation results on some datasets are missing in the original paper.
The highest number for each dataset is in bold,
and the second highest is underlined.
$\dagger$: we report the LaPraDor ~\citep{xu2022laprador} results without ensembling with BM25.
$*$: reproduction with the released checkpoint.}
\label{tab:beir_unsup_results}
\scalebox{0.85}{\begin{tabular}{lcccccccc}
\hline
~ & BM25 & SimCSE & LaPraDor$^\dagger$ & Contriever & cpt$_{\text{300M}}$ & E5-PT$_\text{small}$ & E5-PT$_{\text{base}}$ & E5-PT$_{\text{large}}$ \\ \hline
MS MARCO & 22.8 & 9.4 & 16.9$^*$ & 20.6 & 19.9 & 25.4 & \underline{26.0} & \textbf{26.2} \\
Trec-Covid & \textbf{65.6} & 26.2 & 22.7 & 27.4 & 52.9 & 52.0 & 61.0 & \underline{61.8} \\
NFCorpus & 32.5 & 9.9 & 31.1 & 31.7 & 32.0 &  29.3  & \textbf{35.8} & \underline{33.7} \\
NQ & 32.9 & 11.7 & 18.1 & 25.4 & - &  37.3  & \underline{39.0} & \textbf{41.7} \\
HotpotQA & \textbf{60.3} & 19.8 & 30.3 & 48.1 & 51.5 &  46.0  & \underline{52.4} & 52.2 \\
FiQA & 23.6 & 9.8 & 20.3 & 24.5 & 34.1 &  38.3  & \underline{40.0} & \textbf{43.2} \\
ArguAna & 31.5 & 38.3 & \textbf{45.9} & 37.9 & 38.7 &  42.5  & 42.2 & \underline{44.4} \\
Touche-2020 & \textbf{36.7} & 8.9 & 9.4 & 19.3 & \underline{21.0} & 19.9  & 16.9 & 19.8 \\
CQADupStack & 29.9 & 13.2 & 22.0 & 28.4 & - & 35.0  & \underline{35.4} & \textbf{38.9} \\
Quora & 78.9 & 78.0 & 78.7 & 83.5 & 68.1 &  \underline{85.8}  & 85.7 & \textbf{86.1} \\
DBPedia & 31.3 & 15.0 & 25.0 & 29.2 & 27.2 & 34.5  & \underline{35.4} & \textbf{37.1} \\
Scidocs & 15.8 & 5.5 & 13.3 & 14.9 & - &  19.9  & \underline{21.1} & \textbf{21.8} \\
Fever & \textbf{75.3} & 21.1 & 36.8 & 68.2 & 57.1 &  62.5  & 63.4 & \underline{68.6} \\
Climate-Fever & \textbf{21.3} & 11.8 & 13.8 & 15.5 & \underline{15.8} & 14.5 & 15.4 & 15.7 \\
Scifact & 66.5 & 25.7 & 55.5 & 64.9 & 65.4 & 68.5  & \textbf{73.7} & \underline{72.3} \\ \hline
Average & 41.7 & 20.3 & 29.3 & 36.0 & - & 40.8  & \underline{42.9} & \textbf{44.2} \\
Best on & \underline{5} & 0 & 1 & 0 & 0 & 0 & 2 & \textbf{7} \\ \hline
\end{tabular}}
\end{table}

\noindent
\textbf{Results with Unsupervised Methods }
In Table ~\ref{tab:beir_unsup_results},
we show model results that do not use any labeled data.
When averaged over all $15$ datasets,
E5-PT$_\text{base}$ outperforms the classic BM25 algorithm by $1.2$ points.
To the best of our knowledge,
this is the first reported result that
an unsupervised model can beat BM25 on the BEIR benchmark.
When scaling up to E5-PT$_{\text{large}}$,
we see further benefits from $42.9$ to $44.2$.

In terms of pre-training tasks,
Contriever adopts random cropping,
while LaPraDor combines ICT and dropout-as-positive-instance from SimCSE.
The methods can easily obtain large-scale training data,
while our approach requires more effort in dataset curation.
Such efforts pay off with better results.
Recent studies ~\citep{Lee2022DeduplicatingTD,wenzek2019ccnet,Gao2021ThePA}
also show that improving data quality is a vital step for training large language models.

\begin{table}[ht]
\centering
\caption{Supervised fine-tuning results on the BEIR benchmark.
Results for ANCE ~\citep{xiong2020approximate}, ColBERT ~\citep{Khattab2020ColBERTEA}
and Contriever come from ~\citet{Izacard2021TowardsUD}.
The best result is in bold, and the second best is underlined.}
\label{tab:beir_ft_results}
\scalebox{0.85}{\begin{tabular}{lccccccccc}
\hline
~ & ANCE & GTR$_{\text{base}}$ & ColBERT & Contriever & cpt$_{\text{300M}}$ & GTR$_{\text{large}}$ & E5$_\text{small}$ & E5$_{\text{base}}$ & E5$_{\text{large}}$ \\ \hline
MS MARCO & 38.8 & 42.0 & 40.1 & 40.7 & - & 43.0 & 42.3 & \underline{43.1} & \textbf{44.1} \\
Trec-Covid & 65.4 & 53.9 & 67.7 & 59.6 & 67.9 & 55.7 & 76.8  & \textbf{79.6} & \underline{78.3} \\
NFCorpus & 23.7 & 30.8 & 30.5 & 32.8 & 33.2 & 32.9 & 33.9  & \textbf{36.6} & \underline{36.1} \\
NQ & 44.6 & 49.5 & 52.4 & 49.8 & - & 54.7 & 58.7  & \underline{60.0} & \textbf{62.9} \\
HotpotQA & 45.6 & 53.5 & 59.3 & \textbf{63.8} & 59.4 & 57.9 & 56.3  & 62.2 & \underline{63.3} \\
FiQA & 29.5 & 34.9 & 31.7 & 32.9 & 38.4 & \textbf{42.4} & 34.8 & 36.4 & \underline{38.6} \\
ArguAna & 41.5 & 51.1 & 23.3 & 44.6 & 47.0 & \textbf{52.5} &  46.7  & \underline{51.4} & 49.4 \\
Touche-2020 & 24.0 & 20.5 & 20.2 & 23.0 & \textbf{28.5} & 21.9 & 26.8  & \underline{28.3} & 27.2 \\
CQADupStack & 29.6 & 35.7 & 35.0 & 34.5 & - & 38.4 & 36.1 & \underline{38.9} & \textbf{39.4} \\
Quora & 85.2 & 88.1 & 85.4 & 86.5 & 70.6 & \textbf{89.0} & 87.7 & 87.9 & \underline{88.2} \\
DBPedia & 28.1 & 34.7 & 39.2 & \underline{41.3} & 36.2 & 39.1 & 38.6  & 41.0 & \textbf{42.4} \\
Scidocs & 12.2 & 14.9 & 14.5 & 16.5 & - & 15.8 & 16.4  & \underline{19.0} & \textbf{20.1} \\
Fever & 66.9 & 66.0 & \textbf{77.1} & \underline{75.8} & 72.1 & 71.2 & 53.5 & 58.2 & 65.0 \\
Climate-Fever & 19.8 & \underline{24.1} & 18.4 & 23.7 & 18.5 & \textbf{26.2} & 15.8 & 15.4 & 22.4 \\
Scifact & 50.7 & 60.0 & 67.1 & 67.7 & 67.2 & 63.9 & 65.6  & \textbf{73.1} & \underline{72.6} \\ \hline
Average & 40.5 & 44.0 & 44.4 & 46.6 & - & 47.0 & 46.0  & \underline{48.7} & \textbf{50.0} \\
Best on & 0 & 0 & 1 & 1 & 1 & \underline{4} & 0 & 3 & \textbf{5} \\ \hline
\end{tabular}}
\end{table}

\noindent
\textbf{Results with Supervised Fine-tuning }
In Table ~\ref{tab:beir_ft_results},
we fine-tune our models on supervised datasets and then transfer them to the BEIR benchmark.
Since our fine-tuning datasets include MS-MARCO and NQ,
the corresponding numbers are in-domain results.
For other datasets,
these are zero-shot transfer results.
Our E5$_\text{base}$ model achieves an average nDCG@$10$ of $48.7$,
already surpassing existing methods with more parameters such as GTR$_\text{large}$ ~\citep{Ni2021LargeDE}.
Most datasets benefit from supervised fine-tuning,
but there are also a few exceptions such as FiQA, Scidocs, and Fever, etc.
This is likely due to the lack of enough domain diversity for the fine-tuning datasets.

\subsection{Results on MTEB benchmark}

\begin{table}[ht]
\centering
\caption{Results on the MTEB benchmark ~\citep{Muennighoff2022MTEBMT} (56 datasets in English subset).
Here we only report averaged numbers on each task category for space reasons,
please check out Appendix ~\ref{sec:app_impl} for a detailed version.
BERT-FT$_\text{base}$ uses the same fine-tuning data as E5 but initializes from BERT$_\text{base}$.}
\label{tab:mteb}
\scalebox{0.95}{\begin{tabular}{lcccccccc}
\hline
\multicolumn{1}{l}{\multirow{2}{*}{\# of datasets $\rightarrow$}} & Class. & Clust. & PairClass. & Rerank & Retr. & STS & Summ. & Avg \\
\multicolumn{1}{l}{}                                & 12     & 11     & 3          & 4      & 15    & 10  & 1    & 56  \\ \hline
\multicolumn{9}{l}{\emph{Unsupervised models}}    \\ \hline
\multicolumn{1}{l}{Glove}   &  57.3   &  27.7 & 70.9  & 43.3  & 21.6 & 61.9 & 28.9  & 42.0 \\
\multicolumn{1}{l}{BERT}    &   61.7  & 30.1  &  56.3 & 43.4  &  10.6  &  54.4   &  29.8  &  38.3  \\
\multicolumn{1}{l}{SimCSE-BERT-unsup}   &  62.5  &  29.0  &  70.3   &  46.5  & 20.3  & 74.3 & 31.2  & 45.5 \\
\multicolumn{1}{l}{E5-PT$_\text{small}$}   &    67.0  & 41.7  &  78.2   &  53.1  &  40.8  & 68.8  & \textbf{32.7}  &  54.3  \\
\multicolumn{1}{l}{E5-PT$_\text{base}$}   &    67.9  & \underline{43.4}  &  79.2   &  53.5  &  42.9  & 69.5  & 31.1  &  55.6  \\
\multicolumn{1}{l}{E5-PT$_\text{large}$}  &  69.0  &  \textbf{44.3}  &  80.3   &  54.4  & 44.2  & 69.9 & \underline{32.6}  & 56.6 \\ \hline
\multicolumn{9}{l}{\emph{Supervised models}}  \\ \hline
\multicolumn{1}{l}{SimCSE-BERT-sup}  &  67.3  &  33.4  &  73.7   &  47.5  & 21.8  &  79.1    &  23.3    &  48.7 \\
\multicolumn{1}{l}{BERT-FT$_\text{base}$}  &  68.7  &  33.9  &  82.6  & 50.5  & 41.5  &  79.2  & 29.0  &  55.2  \\
\multicolumn{1}{l}{Contriever}   &   66.7  &  41.1  &  82.5  & 53.1  & 41.9   &  76.5   &  \underline{30.4}    &  56.0   \\
\multicolumn{1}{l}{GTR$_\text{large}$}   &  67.1    &  41.6   &   \underline{85.3}   &  55.4  & 47.4  & 78.2 & 29.5  &  58.3   \\
\multicolumn{1}{l}{Sentence-T5$_\text{large}$}   &  72.3  &  41.7   &  85.0   &  54.0  & 36.7  &  \underline{81.8} &  29.6  & 57.1 \\
\multicolumn{1}{l}{E5$_\text{small}$}     &  71.7 &  39.5  &  85.1  & 54.5  & 46.0  & 80.9  & 31.4 & 58.9 \\
\multicolumn{1}{l}{E5$_\text{base}$}     &  \underline{72.6} &  42.1  &  85.1  & \underline{55.7}  & \underline{48.7}  & 81.0  & 31.0 & \underline{60.4} \\
\multicolumn{1}{l}{E5$_\text{large}$}     &  \textbf{73.1} & 43.3  &  \textbf{85.9}  & \textbf{56.5} & \textbf{50.0}  & \textbf{82.1} & 31.0  & \textbf{61.4}  \\ \hline
\multicolumn{9}{l}{\emph{Larger models}}        \\ \hline
\multicolumn{1}{l}{GTR$_\text{xxl}$}   &  67.4  &  42.4   & 86.1   &  56.7   &  48.5  &  78.4   & 30.6  &  59.0 \\
\multicolumn{1}{l}{Sentence-T5$_\text{xxl}$}   &  73.4   &  43.7  &  85.1    &  56.4  & 42.2  & 82.6  & 30.1  &  59.5 \\ \hline
\end{tabular}}
\end{table}

In Table ~\ref{tab:mteb},
E5 models not only substantially outperform
existing ones with similar sizes,
but also match the results of much larger models.
The top-$2$ models on MTEB leaderboard ~\footnote{\url{https://huggingface.co/spaces/mteb/leaderboard}, as of November 22, 2022}
GTR$_\text{xxl}$ and Sentence-T5$_\text{xxl}$
have $4.8$B parameters,
while our E5$_\text{large}$ model is more than $10\times$ smaller with $300$M parameters.
We expect that our model will benefit from continual scaling up.

Since the difference between BERT-FT$_\text{base}$ and E5$_\text{base}$
is that BERT-FT$_\text{base}$ only has fine-tuning stage,
their performance gap demonstrates the usefulness of contrastive pre-training
on our proposed CCPairs dataset.
For most task categories except Clustering,
performance improves after supervised fine-tuning.
Consistent with prior works ~\citep{Ni2021LargeDE,ni2022sentence},
this once again demonstrates the importance of incorporating human knowledge for learning better text embeddings.
It remains an open question
whether state-of-the-art embeddings can be obtained in a purely self-supervised manner.

\begin{table}[ht]
\centering
\caption{Zero-shot text classification results.
``Majority'' always predicts the majority class label.
Zero-shot BERT$_\text{base}$ uses the average pooling of the last layer as text embeddings.}
\label{tab:zero_shot_tc}
\scalebox{0.95}{\begin{tabular}{lcccccccc}
\hline
\multirow{2}{*}{} & \multicolumn{5}{c}{Zero-shot} & & \multicolumn{2}{c}{Full Fine-tune} \\ \cline{2-6} \cline{8-9}
 & Majority & BERT$_\text{base}$ & E5$_\text{small}$  & E5$_\text{base}$ & E5$_\text{large}$ &  & BERT$_\text{base}$ &  BERT$_\text{large}$  \\ \hline
SST-2 ~\citep{Socher2013RecursiveDM} &  50.9  & 58.9 &  79.7  & 81.3 & \textbf{85.3} & & 93.5 & \textbf{94.9} \\ \hline
\end{tabular}}
\end{table}

Table ~\ref{tab:zero_shot_tc} shows the zero-shot text classification results
on the dev set of the SST-2 dataset ~\citep{Socher2013RecursiveDM}.
By formulating text classification as embedding matching between input and label texts,
our model can be much better than the ``majority'' baseline in a zero-shot setting.
We use the prompt template from Section ~\ref{sec:transfer}.

\subsection{Analysis}

In this section,
we conduct a series of analyses to examine various design choices.
All the numbers in this section are from base-size models.
For the BEIR benchmark,
we choose $6$ datasets with more stable results across different runs.
Some negative results are also listed in Appendix ~\ref{sec:app_neg}.

\begin{table}[ht]
\centering
\caption{Impacts of different batch sizes for contrastive pre-training.}
\label{tab:ablation_batch_size}
\begin{tabular}{lccccccc}
\hline
batch size & NFCorpus & NQ & FiQA & Quora & DBPedia & Scifact & Avg \\ \hline
32k  &  \textbf{35.8}  &  \textbf{39.0}  &  \textbf{40.0}  &  \textbf{85.7}  &  \textbf{35.4}  &  \textbf{73.7}  & \textbf{51.6} \\
8k   &  33.3  & 38.5  &  37.6  &  \textbf{85.7}  & 34.0  &  71.8  &  50.2   \\
1k   &  28.2 &  33.1  &  30.4  &  84.0  &  30.1  & 69.1  &  45.8  \\ \hline
\end{tabular}
\end{table}

\noindent
\textbf{Impacts of Batch Size }
Since we use in-batch negatives for contrastive pre-training,
larger batch size will provide more negatives
and therefore improve the quality of the learned text embeddings.
In Table ~\ref{tab:ablation_batch_size},
increasing batch size from $1$K to $32$K leads to consistent gains
across all $6$ datasets.
It is also possible to train with smaller batch sizes by adding hard negatives ~\citep{ren2021rocketqav2}.
However,
the engineering efforts of mining hard negatives for large datasets (>$100$M)
are non-trivial.

\begin{table}[ht]
\centering
\caption{Fine-tuning with different combinations of labeled data.}
\label{tab:ablation_ft_datasets}
\scalebox{0.95}{\begin{tabular}{lccccc}
\hline
Fine-tuned on   & Retrieval & STS & Classification  & Summ. & MTEB Avg \\ \hline
No fine-tuning &  42.9  &  69.5  &  67.9  &  31.1  &  55.6   \\
MS-MARCO + NQ &  \textbf{50.3}  &  78.3  &  68.3  &  30.6  &  59.0   \\
NLI           &  38.3  &  \textbf{81.1}  &  72.6  & \textbf{31.6}  &  57.3  \\
All above     &  48.7 &  81.0  & \textbf{73.1}  & 31.0  &  \textbf{60.4} \\ \hline
\end{tabular}}
\end{table}

\noindent
\textbf{Fine-tuning Datasets }
GTR models are fine-tuned with ``MS-MARCO + NQ'',
while Sentence-T5 models use NLI instead.
In Table ~\ref{tab:ablation_ft_datasets},
we can see that the ``MS-MARCO + NQ'' setting
performs best on retrieval tasks,
and the NLI data is beneficial for STS and linear probing classification.
Similar observations are also made by ~\citet{Muennighoff2022MTEBMT}.
Combining all of them leads to the best overall scores on the MTEB benchmark.
This also illustrates the importance of dataset diversity for learning text embeddings.

\begin{table}[ht]
\centering
\caption{Data filtering.
For the top $2$ rows,
we train with $1$M random text pairs.}
\label{tab:ablation_data_filter}
\scalebox{0.95}{\begin{tabular}{llccccccc}
\hline
\# of pairs          &            & NFCorpus & NQ & FiQA & Quora & DBPedia & Scifact & Avg  \\ \hline
\multirow{2}{*}{1M}  & w/o filter &  23.0  &  15.1  & 18.5  &  83.1  &  18.2 &  51.4  &  34.9 \\
                     & w/ filter  &  26.8  &  22.7  &  24.5  &  85.0 & 27.5 &  57.5  & \textbf{40.7} \\ \hline
\multirow{2}{*}{All} & w/o filter &  34.5 &  35.4 & 39.1 & 85.7 & 32.9 & 72.5 & 50.0 \\
                     & w/ filter  &  35.8 &  39.0 & 40.0 & 85.7 & 35.4 & 73.7 & \textbf{51.6} \\ \hline
\end{tabular}}
\end{table}

\noindent
\textbf{Data Filtering }
One crucial step in our dataset curation pipeline is filtering out low-quality text pairs.
In Table ~\ref{tab:ablation_data_filter},
when training with $1$M pairs,
using filtered data has a nearly $6$ points advantage.
When all the text pairs are used,
the ``w/o filter'' setting has about $4\times$ more data but
is still behind by $1.6$ points.
Though recent studies ~\citep{Jia2021ScalingUV,Radford2021LearningTV}
show that deep learning models are quite robust to dataset noises,
data filtering still has benefits in improving training efficiency and model quality.

\begin{table}[ht]
\centering
\caption{Comparison of different negative sampling strategies.}
\label{tab:ablation_negatives}
\scalebox{0.95}{\begin{tabular}{lcccccccc}
\hline
            & \# negatives &  NFCorpus & NQ & FiQA & Quora & DBPedia & Scifact & Avg  \\ \hline
In batch    & 32k    &  35.8 &  39.0 & 40.0 & 85.7 & 35.4 & 73.7 & \textbf{51.6} \\
+ pre-batch & 64k    &  29.4  & 27.2 & 29.4 &  84.6  & 25.0  & 64.3  & 43.3  \\
MoCo        & 130k   &  29.7  &  36.1  &  32.0  &  81.6  &  29.9  &  63.6  &  45.5 \\ \hline
\end{tabular}}
\end{table}

\noindent
\textbf{Negative Sampling }
We explore two alternative methods to enlarge the number of negatives:
Pre-batch negatives ~\citep{Lee2021LearningDR} reuse embeddings from previous batches as additional negatives,
while MoCo ~\citep{He2020MomentumCF} introduces a momentum encoder
and uses a FIFO queue to store negatives.
For both approaches,
the negative size can be easily scaled up without incurring much GPU memory overhead.
The downside is that most negatives are produced by an older version of model parameters.
In Table ~\ref{tab:ablation_negatives},
in-batch negatives still perform favorably.
Empirically,
we find that MoCo is more sensitive to certain hyperparameters such as temperature,
better results are possible with more tuning.

\noindent
\textbf{BM25 vs Dense Retrieval }
With the rapid development of dense retrieval models,
can we replace the long-standing BM25 algorithm from now on?
The answer is likely ``\emph{not yet}''.
BM25 still holds obvious advantages in terms of simplicity, efficiency, and interpretability.
For long-tail domains such as Trec-Covid ~\citep{voorhees2021trec}
and retrieval tasks that involve long documents (Touche-2020) ~\citep{bondarenko2022overview}
or rely heavily on exact lexical match (Fever) ~\citep{Thorne2018FEVERAL},
further research efforts are still necessary to improve current dense retrievers.

\section{Conclusion}
In this work, we train a general-purpose text embedding model E5 from weak supervision signals.
We adopt a simple contrastive training framework with in-batch negatives
and learn from a large-scale text pair dataset we harvest from heterogeneous data sources across the web.
E5 offers strong off-the-shelf performance for a wide range of tasks requiring single-vector text representations such as retrieval, semantic textual similarity, and text matching.
When further customized for downstream tasks, E5 achieves superior fine-tuned performance
compared to existing embedding models with $40\times$ more parameters on the large, 56-task MTEB benchmark datasets.

\bibliography{custom}
\bibliographystyle{plainnat}

\appendix

\section{Dataset Details} \label{sec:app_data_details}
For Common Crawl,
we download the 2022-33 snapshot
and cc\_net ~\footnote{\url{https://github.com/facebookresearch/cc_net}} is used for preprocessing
including language identification, de-duplication, language model filtering, etc.
Web pages from the MS-MARCO document ranking corpus are also included.
For the data filtering step,
we examine each pair of passages within a web page instead of just using the title as a query.
For Wikipedia,
we use the version released by ~\citet{Petroni2020KILTAB}.
To avoid possible data contamination,
we remove text pairs that occur in the evaluation datasets based on exact string match.

Reddit data is collected from the year 2018 to August 2022.
For the S2ORC data,
we use a sample weight of $0.3$ during training to avoid over-fitting the scientific domains.

\begin{table}[ht]
\centering
\caption{Details for each data source after filtering.
The ``Others'' category includes ``SimpleWiki'', ``GooAQ'', ``WikiHow'', ``Yahoo Answers''
from \url{https://huggingface.co/datasets/sentence-transformers/embedding-training-data}.}
\label{tab:appendix_datasets}
\scalebox{0.82}{\begin{tabular}{llll}
\hline
data source & type of text pairs                             & random example                                                                                                                                                        & \# of pairs \\ \hline
Wikipedia   & \begin{tabular}[c]{@{}l@{}}(entity+section title, passage)\end{tabular} & \begin{tabular}[c]{@{}l@{}}\textbf{q}: Lexden History\\ \textbf{p}: The site on which Lexden now stands was crossed \\ by the fortifications of iron age Colchester\ldots\end{tabular} & $24$M  \\ \hline
Reddit  & (post, upvoted comment)   & \begin{tabular}[c]{@{}l@{}}\textbf{q}: What makes a client good quality to you?\\ I’m putting together my ideal client \ldots \\ \textbf{p}: Respectful of schedules. And pays on time.\ldots \end{tabular} &  $60$M   \\ \hline
Common Crawl     & (title, passage)   & \begin{tabular}[c]{@{}l@{}}\textbf{q}: Central Intake Unit | Broome County\\ \textbf{p}: Caseworkers from Central Intake assess the \\household and risk of placement. If eligible\ldots \end{tabular} & $69$M \\ \hline
Stackexchange  &  \begin{tabular}[c]{@{}l@{}}(title, answer)\\ (title+description, answer)\end{tabular}  & \begin{tabular}[c]{@{}l@{}}\textbf{q}: Will killing Python made problems for Apache\\ \textbf{p}: Python and Apache aren't related, unless your \\ app is making use of Python. \ldots\end{tabular}    &  $19$M   \\ \hline
S2ORC &  \begin{tabular}[c]{@{}l@{}}(title, abstract)\\ (title, citation title)\\ (abstract, citation abstract)\end{tabular} & \begin{tabular}[c]{@{}l@{}}\textbf{q}: Constructive Dual DP for Reservoir Optimization\\ \textbf{p}: Dynamic programming (DP) is a well established \\ technique for optimization of reservoir manage\ldots \end{tabular} &  $90$M \\ \hline
News  &   \begin{tabular}[c]{@{}l@{}}(title, passage)\\ (highlight, passage)\end{tabular}  & \begin{tabular}[c]{@{}l@{}}\textbf{q}: LG Display reports Q1 operating loss as\ldots \\ \textbf{p}: April 25 (Reuters) - South Korea’s LG Display \\ Co Ltd reported its first quarterly operating loss\ldots \end{tabular}   &  $3$M  \\ \hline
Others   &  misc.   &   misc.    &   $6$M   \\ \hline
All above  &    -   &  -   &  $\sim 270$M  \\ \hline
\end{tabular}}
\end{table}

For the BEIR benchmark,
we use the 15 datasets that provide public downloads:
MS MARCO ~\citep{Campos2016MSMA}, Trec-Covid ~\citep{voorhees2021trec}, NFCorpus ~\citep{boteva2016full},
NQ ~\citep{Kwiatkowski2019NaturalQA}, HotpotQA ~\citep{yang2018hotpotqa}, FiQA ~\citep{maia201818}, ArguAna ~\citep{wachsmuth2018retrieval},
Touche-2020 ~\citep{bondarenko2022overview}, CQADupStack ~\citep{hoogeveen2015cqadupstack}, Quora, DBPedia ~\citep{hasibi2017dbpedia},
Scidocs ~\citep{cohan2020specter}, Fever ~\citep{Thorne2018FEVERAL},
Climate-Fever ~\citep{diggelmann2020climate}, and Scifact ~\citep{wadden2020fact}.

\section{Implementation Details} \label{sec:app_impl}
We list the hyperparameters in Table ~\ref{tab:appendix_hyper}.
Since some evaluation datasets have long texts,
we freeze the position embeddings during both pre-training and fine-tuning
and set the maximum text length to $512$ for evaluation.

For the Quora duplicate retrieval task in the BEIR benchmark,
we add prefix ``\emph{query: }'' to all the questions.
For other retrieval tasks,
we use ``\emph{query: }'' and ``\emph{passage: }'' prefixes correspondingly.

The MS-MARCO results in Table ~\ref{tab:in_domain} use document titles
provided by RocketQA ~\citep{ren2021rocketqav2}.
This evaluation setup is consistent with most state-of-the-art dense retrievers.
However,
the MS-MARCO data from the BEIR benchmark does not have titles,
so the results are expected to be lower.

\begin{table}[ht]
\centering
\caption{Model configurations.}
\label{tab:app_model_config}
\begin{tabular}{lccc}
\hline
      & \# layers & hidden size & \# params \\ \hline
E5$_\text{small}$ &  $12$    &  $384$    & $33$M    \\
E5$_\text{base}$  &  $12$  &  $768$     &   $110$M    \\
E5$_\text{large}$ &  $24$   &  $1024$  &   $330$M   \\ \hline
\end{tabular}
\end{table}

\begin{table}[ht]
\centering
\caption{Hyperparameters for contrastive pre-training and fine-tuning.}
\label{tab:appendix_hyper}
\scalebox{0.9}{\begin{tabular}{lccccccc}
\hline
 & \multicolumn{3}{c}{pre-training} & & \multicolumn{3}{c}{fine-tuning} \\ \cline{2-4} \cline{6-8}
 & E5-PT$_\text{small}$  &  E5-PT$_\text{base}$  & E5-PT$_\text{large}$ &   & E5$_\text{small}$   & E5$_\text{base}$    & E5$_\text{large}$   \\ \hline
learning rate & $3 \times 10^{-4}$ &  $2 \times 10^{-4}$  &   $10^{-4}$  &  &  $3\times 10^{-5}$   &  $2\times 10^{-5}$  &  $10^{-5}$ \\
GPUs &   $16$   &   $32$  &  $64$  &  &  $8$  &  $8$  &  $8$  \\
warmup steps &  $1000$  &  $1000$    &   $1000$  &  &  $400$  &  $400$      &  $400$   \\
batch size &   $32$K  &   $32$K  &   $32$K  &  &  $256$  &  $256$  &  $256$ \\
max steps  &  $20$K &  $20$K   &  $20$K  &  & n.a.  & n.a.   &  n.a.    \\
max length  &  $128$ &  $128$   &  $128$  &  & $192$  & $192$  &  $192$ \\
epochs &   n.a.   &   n.a.   &  n.a.  &  &  $3$  &  $3$      &  $3$    \\
$\tau$ &   $0.01$  &   $0.01$     &  $0.01$  &  &    $0.01$  &    $0.01$  &    $0.01$  \\
$\alpha$ &   n.a.  &   n.a.     &  n.a.  &  &  $0.2$  &    $0.2$  &    $0.2$  \\
weight decay &    $0.01$  &    $0.01$  &  $0.01$ &  &  $0.01$   &  $0.01$   &  $0.01$    \\
hard negatives  &  $0$ &  $0$  &  $0$  &   &  $7$   &   $7$  &   $7$    \\ \hline
\end{tabular}}
\end{table}

\noindent
\textbf{In-domain Evaluation }
We report results for in-domain datasets in Table ~\ref{tab:in_domain}.
These results can help illustrate the benefits brought by contrastive pre-training
when abundant in-domain labeled data are available.
For MS-MARCO passage ranking,
MRR@10 and Recall@1k are reported.
For the NQ dataset,
Recall@20 and Recall@100 are the main metrics.

\begin{table}[ht]
\centering
\caption{In-domain results.
``target pre-train'' refers to intermediate pre-training on the target corpus before supervised fine-tuning.
For NQ, we use the passage retrieval setting from DPR ~\citep{Karpukhin2020DensePR}.}
\label{tab:in_domain}
\begin{tabular}{lcccccc}
\hline
 &  \multirow{2}{*}{\begin{tabular}[c]{@{}l@{}}target pre-train?\end{tabular}}  & \multicolumn{2}{c}{MS-MARCO} & & \multicolumn{2}{c}{NQ} \\ \cline{3-4} \cline{6-7}
 &   &    MRR@10     &     R@1k & &  R@20     &   R@100   \\ \hline
ANCE ~\citep{xiong2020approximate}  &  \xmark  &     33.0    &  95.9   &  &   81.9     &  87.5   \\
RocketQAv2 ~\citep{ren2021rocketqav2}  & \xmark   &  38.8 &  98.1  &  &  83.7 &  89.0   \\
SimLM ~\citep{Wang2022SimLMPW} & \cmark   &   \textbf{41.1}    &  \textbf{98.7}   &  &   85.2  &  89.7  \\ \hline
E5$_\text{small}$ &  \xmark  &  37.5    &  98.1    &  &  84.6   &  89.8  \\
E5$_\text{base}$ &  \xmark  &  38.5    &  98.5    &  &  86.1   &  \textbf{90.7}  \\
E5$_\text{large}$ &  \xmark   &   39.4   &  \textbf{98.7}   &  &   \textbf{86.4}   &  90.5 \\ \hline
\end{tabular}
\end{table}

\section{Negative Results} \label{sec:app_neg}
    Here are some attempts that we eventually give up on:
    
    \noindent
    \textbf{Adding BM25 hard negatives }
    Similar to DPR ~\citep{Karpukhin2020DensePR},
    we add one BM25 hard negative for each positive pair during training.
    When using $15$M data,
    this strategy improves the overall results by $\sim 0.5$ points on the BEIR benchmark.
    However,
    running the BM25 algorithm over a $250$M+ dataset is too time-consuming even with multi-node and multi-process parallelism.
    
    \noindent
    \textbf{Using RoBERTa instead of BERT for initialization }
    Though RoBERTa shows consistent gains on many NLP tasks,
    we empirically find that RoBERTa performs worse than BERT initialization
    on most of the BEIR benchmark datasets.
    
    \noindent
    \textbf{Auxiliary MLM objective }
    We add a masked language modeling loss for $25$\% of the training text pairs.
    The numbers are on par with removing this auxiliary objective,
    but the training cost goes up.

\begin{table}[ht]
\centering
\caption{Results for each dataset in the MTEB benchmark ~\citep{Muennighoff2022MTEBMT}.
The numbers for the Retrieval category are not included here
since the datasets are the same as the BEIR benchmark.}
\label{tab:appendix_mteb}
\scalebox{0.9}{\begin{tabular}{lccccccc}
\hline
 & \multicolumn{3}{c}{unsupervised} & & \multicolumn{3}{c}{supervised} \\ \cline{2-4} \cline{6-8}
 & E5-PT$_\text{small}$  &  E5-PT$_\text{base}$  & E5-PT$_\text{large}$ &  & E5$_\text{small}$   & E5$_\text{base}$    & E5$_\text{large}$   \\ \hline
AmazonCounterfactualClassification & 71.7 & 73.6 & 70.4 & & 76.2 & 79.7 & 77.7 \\
AmazonPolarityClassification & 76.1 & 77.0 & 83.2 & & 87.5 & 88.0 & 90.1 \\
AmazonReviewsClassification & 35.0 & 35.8 & 37.4 & & 42.6 & 42.7 & 43.0 \\
Banking77Classification & 82.1 & 82.9 & 83.5 & & 81.9 & 83.3 & 84.1 \\
EmotionClassification & 42.2 & 44.2 & 43.5 & & 46.9 & 49.4 & 48.1 \\
ImdbClassification & 67.9 & 67.3 & 77.7 & & 75.6 & 76.0 & 82.1 \\
MassiveIntentClassification & 70.2 & 71.1 & 70.8 & & 72.2 & 72.3 & 73.2 \\
MassiveScenarioClassification & 74.6 & 75.4 & 75.9 & & 75.8 & 76.8 & 77.4 \\
MTOPDomainClassification & 91.3 & 92.3 & 93.2 & & 92.1 & 93.2 & 93.9 \\
MTOPIntentClassification & 71.9 & 74.0 & 74.2 & & 73.2 & 74.8 & 76.4 \\
ToxicConversationsClassification & 67.0 & 67.4 & 66.1 & & 72.8 & 74.1 & 70.6 \\
TweetSentimentExtractionClass. & 54.4 & 53.3 & 52.5 & & 63.3 & 61.4 & 61.2 \\ \hline
ArxivClusteringP2P & 47.9 & 49.3 & 49.4 & & 44.1 & 44.6 & 46.2 \\
ArxivClusteringS2S & 39.9 & 42.8 & 43.6 & & 37.1 & 40.5 & 41.4 \\
BiorxivClusteringP2P & 38.5 & 38.8 & 39.2 & & 35.8 & 36.2 & 37.6 \\
BiorxivClusteringS2S & 35.4 & 36.5 & 36.7 & & 31.9 & 32.7 & 35.1 \\
MedrxivClusteringP2P & 34.4 & 33.7 & 33.3 & & 31.3 & 31.5 & 32.3 \\
MedrxivClusteringS2S & 32.0 & 32.1 & 32.2 & & 28.2 & 28.3 & 29.7 \\
RedditClustering & 46.9 & 49.3 & 52.4 & & 42.9 & 48.2 & 50.7 \\
RedditClusteringP2P & 60.2 & 64.4 & 64.6 & & 56.4 & 62.2 & 61.4 \\
StackExchangeClustering & 57.7 & 60.2 & 63.3 & & 59.1 & 63.9 & 65.0 \\
StackExchangeClusteringP2P & 32.0 & 34.0 & 34.7 & & 30.3 & 32.6 & 33.6 \\
TwentyNewsgroupsClustering & 34.4 & 36.2 & 37.9 & & 37.5 & 42.6 & 43.8 \\ \hline
SprintDuplicateQuestions & 91.6 & 90.8 & 92.0 & & 95.3 & 94.9 & 95.4 \\
TwitterSemEval2015 & 60.0 & 62.8 & 64.7 & & 74.2 & 74.4 & 76.1 \\
TwitterURLCorpus & 83.2 & 84.0 & 84.1 & & 85.8 & 86.0 & 86.3 \\ \hline
AskUbuntuDupQuestions & 57.8 & 57.6 & 58.3 & & 59.4 & 59.7 & 60.1 \\
MindSmallReranking & 29.0 & 29.6 & 29.2 & & 29.6 & 30.1 & 30.8 \\
SciDocsRR & 81.1 & 82.6 & 84.3 & & 79.8 & 82.9 & 83.9 \\
StackOverflowDupQuestions & 44.4 & 44.2 & 45.8 & & 49.1 & 50.1 & 51.3 \\ \hline
BIOSSES & 69.2 & 71.9 & 69.7 & & 84.2 & 85.1 & 84.7 \\
SICK-R & 66.6 & 68.7 & 69.7 & & 78.9 & 79.7 & 80.5 \\
STS12 & 60.7 & 57.9 & 54.7 & & 75.2 & 74.2 & 75.9 \\
STS13 & 71.1 & 73.5 & 74.0 & & 81.8 & 83.3 & 85.2 \\
STS14 & 64.2 & 64.0 & 65.3 & & 78.5 & 78.5 & 80.5 \\
STS15 & 74.3 & 75.4 & 75.8 & & 87.5 & 88.4 & 88.8 \\
STS16 & 76.6 & 79.8 & 80.1 & & 84.6 & 84.2 & 85.3 \\
STS17 & 78.3 & 77.2 & 76.0 & & 87.9 & 87.2 & 89.4 \\
STS22 & 59.2 & 56.2 & 62.8 & & 63.8 & 62.9 & 63.0 \\
STSBenchmark & 67.7 & 70.5 & 70.9 & & 86.4 & 86.2 & 87.2 \\ \hline
SummEval & 32.7 & 31.1 & 32.6 & & 31.4 & 31.0 & 31.0 \\ \hline
\end{tabular}}
\end{table}

\end{document}